\documentclass[a4paper,fleqn]{cas-sc}

\usepackage{tkz-graph}
\usepackage{amssymb}
\usepackage{amsmath}
\usepackage{graphicx}
\usepackage[retainorgcmds]{IEEEtrantools}
\usepackage[colorlinks]{hyperref}
\hypersetup{
  citecolor = {blue}
}
\usepackage{natbib}
\usepackage{siunitx}
\usepackage{tabularx}
\usepackage{geometry}
\usepackage{pdflscape}
\usepackage{amssymb}

\usepackage{enumitem}
\usepackage[normalem]{ulem}
\useunder{\uline}{\ul}{}
\usepackage{setspace} 
\usepackage[utf8]{inputenc}
\usepackage[linesnumbered, ruled, vlined]{algorithm2e}
\usepackage{rotating}
\usepackage{mdframed}
\usepackage{algorithm2e}

\usepackage{nicefrac}
\usepackage{xspace}
\usepackage{float}
\usepackage{caption}
\usepackage{subcaption}
\usepackage{rotating, lscape, longtable, tabu, booktabs, siunitx, multirow}
\usepackage[capitalise, noabbrev]{cleveref}
\usepackage{verbatim}
\usepackage{mathtools}
\usepackage{lipsum}

\usepackage{enumerate}
\usepackage[intoc]{nomencl}
\usepackage{multicol} 
\usepackage{wrapfig} 
\usepackage{framed}  
\usepackage{nomencl}
\def\tsc#1{\csdef{#1}{\textsc{\lowercase{#1}}\xspace}}
\tsc{WGM}
\tsc{QE}
\tsc{EP}
\tsc{PMS}
\tsc{BEC}
\tsc{DE}

\begin{document}
\let\WriteBookmarks\relax
\def\floatpagepagefraction{1}
\def\textpagefraction{.001}

\ExplSyntaxOn
\cs_if_exist:NTF \printemails {} {\cs_new:Npn \printemails {}}
\cs_if_exist:NTF \printurls {} {\cs_new:Npn \printurls {}}
\cs_if_exist:NTF \printorcid {} {\cs_new:Npn \printorcid {}}
\cs_if_exist:NTF \printfacebook {} {\cs_new:Npn \printfacebook {}}
\cs_if_exist:NTF \printtwitt {} {\cs_new:Npn \printtwitt {}}
\ExplSyntaxOff

\shorttitle{Y.Wang and Y.Zhao}

\shortauthors{Y.Wang and Y.Zhao}

\title [mode = title]{Multiple Ships Cooperative Navigation and Collision Avoidance using Multi-agent Reinforcement Learning with Communication}

\author[1]{Yufei Wang}



\credit{Conceptualization, Methodology, Investigation
, Writing - Original Draft}

\affiliation[1]{organization={National University of Singapore},
    city={Singapore},
    postcode={119077}, 
    country={Singapore}}

\author[2]{Yang Zhao}[style=chinese]

\credit{Resources, Supervision, Writing - Review & Editing}

\affiliation[2]{organization={Nanyang Technological University},
    city={Singapore},
    postcode={639798},
    country={Singapore}}
\cormark[1]
\ead{zhao0466@e.ntu.edu.sg}

\begin{abstract}
In the real world, unmanned surface vehicles (USV) often need to coordinate with each other to accomplish specific tasks. However, achieving cooperative control in multi-agent systems is challenging due to issues such as non-stationarity and partial observability. Recent advancements in Multi-Agent Reinforcement Learning (MARL) provide new perspectives to address these challenges. Therefore, we propose using the multi-agent deep deterministic policy gradient (MADDPG) algorithm with communication to address multiple ships' cooperation problems under partial observability. We developed two tasks based on OpenAI's gym environment: cooperative navigation and cooperative collision avoidance. In these tasks, ships must not only learn effective control strategies but also establish communication protocols with other agents. We analyze the impact of external noise on communication, the effect of inter-agent communication on performance, and the communication patterns learned by the agents. The results demonstrate that our proposed framework effectively addresses cooperative navigation and collision avoidance among multiple vessels, significantly outperforming traditional single-agent algorithms. Agents establish a consistent communication protocol, enabling them to compensate for missing information through shared observations and achieve better coordination.
\end{abstract}



\begin{keywords}
Multi-agent reinforcement learning \sep MADDPG \sep Multiple ships control \sep Navigation \sep Collision avoidance \sep Maritime communication 
\end{keywords}

\maketitle

\begin{mdframed}
    \begin{multicols}{2}
        \textbf{Nomenclature}
        \begin{description}[labelwidth=2cm, labelsep=0.2cm, itemindent=0cm, leftmargin=!, itemsep=-0.05 cm]

            \item[\normalfont RL] Reinforcement Learning
            \item[\normalfont USV] Unmanned surface vehicle
            \item[\normalfont MDP] Markov decision process
            \item[\normalfont OS] Own ship
              \item[\normalfont TS] Target ship
              \item[\normalfont DQN] Deep Q-learning
              \item[\normalfont PPO] Proximal policy optimization
              \item[\normalfont DDPG] Deep deterministic policy gradient
              \item[\normalfont MARL]
              Multi-agent reinforcement learning
              \item [\normalfont MADDPG] Multi-agent deep deterministic policy gradient
              \item [\normalfont CTDE] Centralized training with decentralized execution
              \item [\normalfont MPE]
              Multi-Agent Particle Environment
              \item [\normalfont Dec-POMDP] Decentralized partially observable Markov decision process
              \item[\normalfont MMG] Mathematic Model Groups
              \item[$\cal I$] A set of $n$ agents
              \item[$\cal S$] A set of state $\boldsymbol {s}$ 
              \item[$\boldsymbol{\cal A}$] The finite set of joint actions
              \item [$\boldsymbol{\Omega}$]
              The finite set of joint observations      \item[$\cal T$] The set of transition functions
              \item[$\cal R$] The set of immediate reward functions
              \item[$\boldsymbol{a}$] A set of actions
              \item[$\cal O$] The set of observation probability function
              \item [$\gamma$] Discount factor
              \item [$\boldsymbol{o}$] A set of individual observations
              \item [$\mu_i$] Actor (policy) network of agent $i$
              \item [$Q_i$] Critic network of agent $i$
              \item [$a_i^t$] The action taken by agent $i$ at time step $t$.
              \item[$o_i^t$] The observation of agent $i$ and time step $t$
              \item[$\theta_i^Q$] The weight of critic network for agent $i$
              \item[$\theta_i^{Q'}$] The weight of target critic network for agent $i$
              \item[$\theta_i^{\mu}$] The weight of actor network for agent $i$   
              \item[$\alpha$] Learning rate
 
        \end{description}
    \end{multicols}
\end{mdframed}

\begin{mdframed}
    \begin{multicols}{2}
        \textbf{Nomenclature}
        \begin{description}[labelwidth=2cm, labelsep=0.2cm, itemindent=0cm, leftmargin=!, itemsep=-0.05 cm]
            \item[$\tau$] Soft update rate
            \item[${\cal W}_t$] Ornstein-Uhlenbeck noise
            \item[$r_t$] The collective reward at time step $t$
            \item[$m_s$] Ship mass
              \item[$m_x$, $m_y$] Ship's added mass on $x$ and $y$ direction.
              \item[$v_m$] Lateral velocity at midship            \item[$X_H$, $Y_H$, $N_H$] Surge force, lateral force, yaw moment around midship by steering
            \item[$X_G$]Longitudinal coordinate of center of gravity of ship
              \item [$X_R$, $Y_R$, $N_R$] Surge force, lateral force, yaw moment around midship by steering
              \item [$r_y$] Yaw rate
              \item [$I_{zG}$]
              Moment of inertia of ship around center of gravity
              \item [$\Psi$] Heading angle
              \item[$\beta$] Drift angle
              \item[$L$] Ship length
              \item[$\delta_d$, $\delta$] Desired rudder angle, and real rudder angle
              \item[$J_z$] Added moment of inertia              \item[$P_e$] Probability of flip error in BSC channel
              \item [$\boldsymbol{m}_i$]
              Message sent by agent $i$.      \item[${\cal N} (0, \sigma^2)$] The Gussian white noise added to the AWGN channel
              \item[$o_0-x_0y_0z_0$] Fixed earth reference frame
              \item[$o-xyz$] Local reference system
              \item[$\cal O$] The set of observation probability function
              \item [$P_e$] Flip error in BSC channel
              \item [$\boldsymbol{m}_i$] communication vectors sent by agent $i$
              \item [$k$] Message length
              \item [$N$] Number of ships
              \item [$M$] Number of landmarks
              \item[$p_x, p_y$] Ship's current position
              \item[$\boldsymbol{P}_{\text{goal}}$] Ship's goal position
              \item[$U_x, U_y$] Ship's velocities in local frame
              \item[$\cal B$] Batch size
              \item[$\cal M$] Total training episode  
              \item[$L_e$] Maximum steps per episode
        \end{description}
    \end{multicols}
\end{mdframed}

\begin{multicols}{2}
\section{Introduction}
USVs are increasingly capturing the interest of maritime operations due to their remarkable flexibility \citep{peng2017development} and cost efficiency \citep{jo2019low}. In real-world applications, USVs frequently need to interact with human operators and other vessels. As illustrated in Figure~\ref{FIG:1}, smart USVs must be able to interpret and act upon commands from human operators \citep{peeters2020inland}, such as transporting cargo from a port to an offshore platform \citep{gu2021autonomous} or navigating to specific locations to function as base stations for wireless communication \citep{wang2022unmanned}. Furthermore, effective collaboration between USVs and both manned and unmanned vessels is crucial for tasks such as collision avoidance \citep{chen2022survey}. Developing USVs that can seamlessly integrate with other entities in complex maritime environments is essential for enhancing safety and boosting economic productivity at sea \citep{tan1993multi}.
\subsection{Literature review}
Due to recent advances in artificial intelligence, many studies modeled ship control problem as a Markov decision process (MDP) and then solve it using reinforcement learning (RL, \cite{sutton2018reinforcement}). \cite{shen2019automatic} studied the automatic collision avoidance of multiple ships in congested water ways by using deep Q-learning (DQN). They use twelve detection lines to measure the distance between the own ship (OS) and target ships (TS), and the latest five records of detected distance are defined as state space. They also incorporate human experience to this model by considering ship bumper, ship domain, and predicted area of danger. The results derived from numerical simulations and experiments show that this method is effective in avoiding collisions in complex environments. \cite{zhao2019colregs} proposed a collision avoidance strategy for OS in multiple ship encounter situations based on proximal policy optimization (PPO) algorithm. They divided the space into four regions according to COLREGs, and only the nearest ship in each region will be treated as TS. In addition to the state information of OS, e.g., course angle error, relative distance to the goal, and angular rates of the heading angle, etc. The states of all TSs like positions, velocities, and ship lengths, were also used as the input of neural network, while the output of the network is discrete rudder angle of the OS.  The OS needs to avoid collisions with TSs in a COLREGs complaint way for receiving best reward. \cite{woo2020collision} designed a hierarchy control architecture. They modeled decision making procedure as a semi-Markov decision process. The input of the DQN network was a grid map representation of OS’s surrounding environment. The output of DQN was high-level behaviors such as path following, starboard avoidance, and port avoidance, while low-level actions such as the adjustment of speed and rudder angle were determined by conventional PID controller. \cite{guo2020autonomous} proposed a method for ship path planning that utilizes the automatic identification systems data. The Deep Deterministic Policy Gradient (DDPG) algorithm was applied to addresses the challenge of continuous space. \cite{chun2021deep} considered collision risks in multiple ship encounter situations by combining the concept of ship domain and the closet point approach. The ship with highest collision risk is selected as TS, and this collision risk is also used as one of the inputs of policy agent. In this case, the calculation of safety indicators directly determines whether this algorithm can generate correct paths, which means OS must perfectly monitor the status of all surrounding ships at any given time. In fact, a small error or uncertainty in measurement of TSs’ states will lead to a huge variance in collision risk assessment \citep{sato1998study}.  \cite{sawada2021automatic} combined PPO with long short-term memory for addressing the challenge of longer safe passing distances in multiple ship collision avoidance. Thus, the model can store and utilize historical information about the environment, including the ship's previous states and actions. This allows the trained model to make more complex decisions, such as giving up collision avoidance and returning to the waypoint in the midst of operation. \cite{xu2022colregs} proposed a hybrid model for creating collision-free paths. They first used a hybrid risk assessment method that combines various kinds of safety indicators to determine whether the ship should take actions to avoid obstacles and to output possible encounter situations.  The information about OS, the relative distance to TSs and encounter situations were used as the input of DDPG algorithm. They also introduced priority sampling technique with cumulative pruning to enhance sampling efficiency. \cite{wang2023cooperative} explored the marine search and rescue operations in a cooperative USV-UAV system. The UAV was used to provide an expansive overhead perspective of the task area, as the sensing devices on the USV, like cameras and radars, have a restricted range from their lower vantage point. The authors introduced a CNN-based visual navigation architecture for positioning USVs and floating objects, along with a RL control strategy for the USV to approach and circle around floating objects. \cite{xu2023real} proposed a real-time planning and collision avoidance control method based on DDPG. 
They employed a layered sampling exploration method to reduce the risk of overfitting and shorten training time compared to traditional random sampling algorithms.
\citep{zheng2024adaptive} employed supervised learning to categorize 2D images into three encounter situations. The extracted data is then fed into a DQN for collision avoidance. By leveraging data augmentation techniques, they expanded the dataset, ensuring that the ship can make right decisions across a wide range of scenarios. 

\subsection{Existing challenges}
Previous studies have successfully applied RL based method to ship control problems, strongly demonstrating its effectiveness and high efficiency. However, there still exists several problems that hinders the real-world application of RL.

To begin with, current methods often assume perfect observation conditions to satisfy MDP framework. However, perfect observation could be unrealistic in real-world maritime environments, even if some ship are equipped with advanced sensors \citep{aldous2015uncertainty,thombre2020sensors, dalheim2021uncertainty}. On the one hand, sensors like cameras, sonars and radar can only detect objects within a specific range and have a fixed field of view, which means there are blind spots in the ship's observation.  Additionally, the resolution of sensors may not be sufficiently high to detect small objects or determine real distance to other USVs \citep{fefilatyev2012detection}. On the other hand, sensors are sensitive to environment conditions. Visibility, weather, and sea state can significantly deteriorate sensor performance. For instance, radar and sonar systems are frequently overwhelmed by `sea clutter' \citep{ward2006sea}, particularly in congested maritime zones, where unwanted echoes from waves, debris, buoys, and land masses can obscure the radar imagery and impede the accurate identification of target signatures. Therefore, develop a ship control strategy that is capable of making decisions based on incomplete or imperfect observation is of vital importance. 

Another gap between existing studies and reality is that multiple ships control problem is still not well explored. In a real-world ocean environment, there always exist multiple ships that must interact, coordinate, compete, or adapt to one another, but conventional RL algorithms designed for single agent cannot be simply extended to such a multi-agent system \citep{busoniu2008comprehensive}. The first reason is non-stationarity \citep{gronauer2022multi}. In a multi-agent system, when agents continuously adapt their strategies to meet their own interests, the environment from the perspective of any single agent becomes dynamic \citep{hernandez2017survey}. Such non-stationarity essentially undermines the Markovian property that guarantees the theoretical convergence of most single-agent methods. The second reason is credit assignment \citep{foerster2018counterfactual,nguyen2018credit, rashid2020monotonic}. Single-agent RL is designed to help one agent learn how to make good decisions based on the rewards it receives from its actions. However, when there are multiple agents involved, all interacting and influencing each other's outcomes. In this situation, single-agent RL may struggle to figure out who should get credit or blame for the results, because the results are the combined effect of everyone's actions.

Third, many communication-related problems are often neglected within the context of RL, although communication plays a crucial role in maritime navigation \citep{grant2009gps,alqurashi2022maritime}. For example, when two ships meet, effective communication allows them to exchange their positions, headings, speeds, and planned maneuvers to prevent collisions. Can these ships learn to develop a shared communication protocol that defines what to say, when to say and how to interpret by themselves? \citep{zhu2022survey} How significantly does noise impact their mutual communication? \citep{tung2021effective} Can they devise a robust strategy that withstands noise interference? Additionally, can mutual communication enhance coordination between ships? We aim to provide deeper insights into these questions.

\begin{figure*}
\includegraphics[scale=0.45]{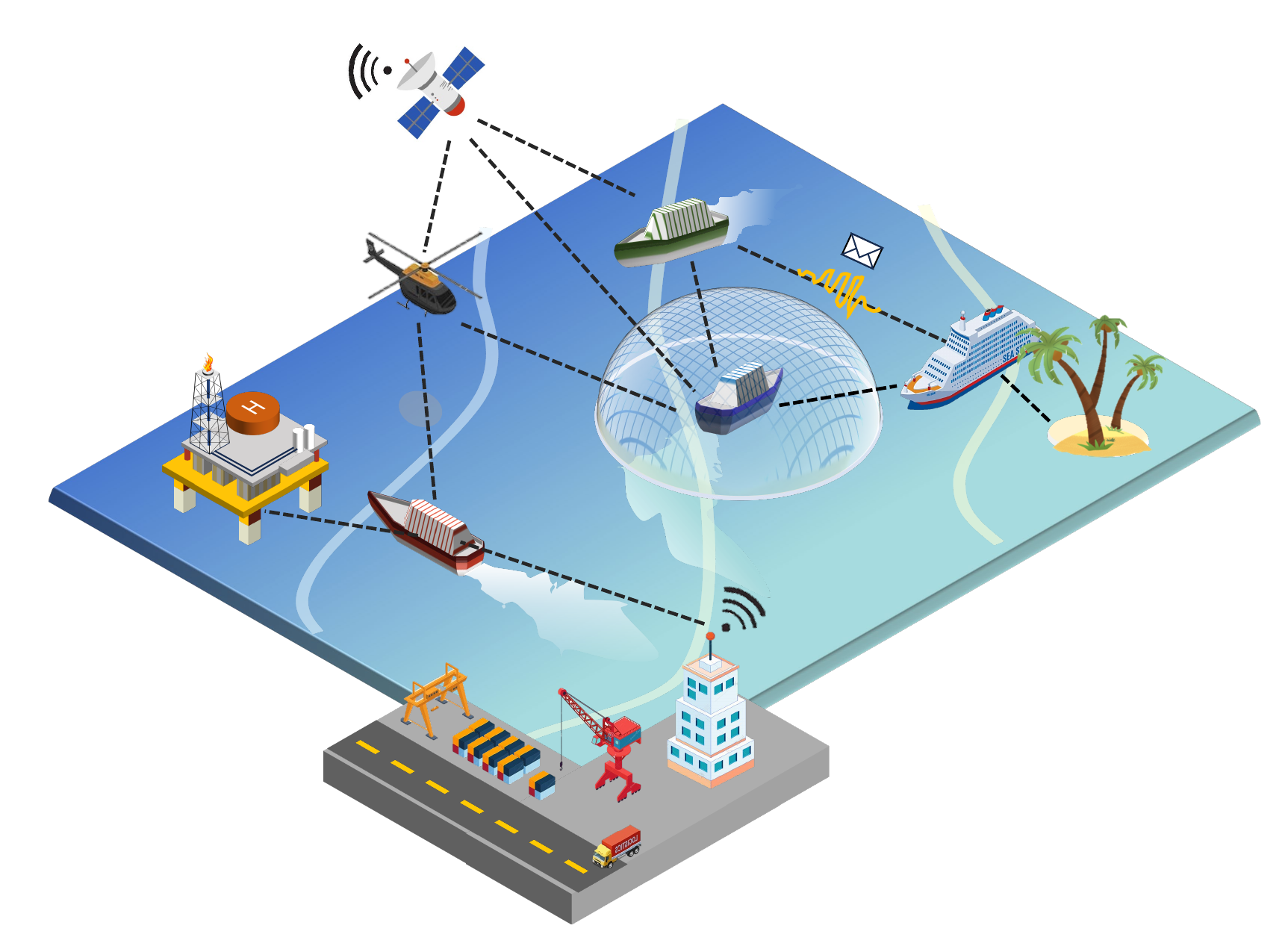}
\caption{Schematic diagram of typical work scenarios of USVs. Due to the technological limitations and environmental uncertainties, USVs can only partially observe the environment most of the time, which means USVs should have the ability to make decisions based on imperfect information. In addition, a truly smart USV should know how to collaborate with other entities to accomplish certain tasks, e.g., cooperative navigation and cooperative collision avoidance}
\label{FIG:1}
\end{figure*}

\subsection{Motivations and contributions}
To address these issues, our study proposes the use of multi-agent reinforcement learning (MARL), particularly the multi-agent deep deterministic policy gradient algorithm (MADDPG, \cite{lowe2017multi}), for collaborative navigation and collision avoidance in partially observable maritime settings. MADDPG operates on a centralized training with decentralized execution (CTDE) framework \citep{oliehoek2008optimal, foerster2016learning}. During training, a central critic utilizes information about the actions, policies, and rewards of all agents, effectively managing non-stationarity by considering the collective actions of all agents. After training, each agent independently executes its policy based on local observations, ensuring scalability and robustness in operation. MADDPG also facilitates agent communication through explicit messages, allowing for a comprehensive understanding of the environment and addressing partial observability.

We evaluate our proposed method using two scenarios adapted from OpenAI's Multi-Agent Particle Environment (MPE, \cite{mordatch2017emergence}). The first scenario, `ships cooperative navigation', involves a human proxy, one or more ships, and several landmarks, where ships must communicate with the human proxy over a noisy channel to identify their target landmarks. The second scenario, `ships cooperative collision avoidance', requires ships to navigate to their destinations while avoiding collisions in a COLREGs-compliant manner. In this scenario, ships can only partially observe their environment and must communicate their information to other ships, considering bandwidth constraints that require concise message transmission. These scenarios test the effectiveness of our proposed MADDPG approach in real-world applications, emphasizing the importance of collaborative communication and decision-making in autonomous maritime navigation.

The main contributions of this research are as follows:
\begin{itemize}
    \item \textbf{Modeling of Maritime Environment as a Multi-Agent System.} We conceptualize the maritime environment as a multi-agent system where multiple agents simultaneously learn strategies. This approach is novel in addressing the complex dynamics of maritime operations that involve multiple ships.
    \item  \textbf{Application of the MADDPG Algorithm.} We propose the use of the MADDPG algorithm for path planning and collision avoidance in scenarios requiring multiple ships' cooperation. MADDPG's framework of centralized training and decentralized execution effectively addresses the challenges of non-stationarity and scalability, which are limitations in single-agent algorithms.
    \item \textbf{Partial Observability and Communication Protocol Development.} Our approach does not assume that Unmanned Surface Vehicles (USVs) have perfect environmental awareness. Instead, we focus on decision making under imperfect or incomplete observations. To enhance coordination under these conditions, we enable agents to communicate through explicit messaging. This innovation requires USVs to develop not only maneuvering skills but also a communication protocol for effective interaction.
    \item \textbf{Incorporation of Real-World Communication Constraints.} We address the often-overlooked aspects of bandwidth and noise limitations in maritime communication. Our research explores the capability of our proposed method to transmit succinct messages under bandwidth constraints and its effectiveness in noisy environments, thereby adding realism to our model.
\end{itemize}
The rest of paper is organized as follows. Section II presents the system model and problem formulation. In Section III, we demonstrate the MADDPG method works well for multiple ships cooperative scenarios. Simulation results and discussions are presented in Section IV. Finally, conclusion of this paper is elaborated in Section V.

\section{Method}
We formulate multi-ship cooperation task as a  decentralized partially observable Markov decision process. Building on this, the MADDPG algorithm is used for decision making. We incorporate a ship dynamics model into the environment to describe ship movement. Additionally, the decision-making process integrates constraints from the International Regulations for Preventing Collisions at Sea to ensure USVs make legal and safe decisions in various scenarios. Lastly, to investigate how noisy channel will affect the coordination among agents, we briefly introduce two prevailing noise models.
\subsection{The Dec-POMDP Model}
Given that perfect global observations are not always guaranteed in real-world ship navigation, and considering the cooperative nature of the task, we model the multi-ship cooperation problem as a decentralized partially observable Markov decision process (Dec-POMDP). A Dec-POMDP \citep{bernstein2002complexity} offers a probabilistic framework to formulate multiagent decision making problems under uncertainty and incomplete information. In a Dec-POMDP setting, each agent must take actions solely based on imperfect local observation and collaborate with each other to maximum a single global reward function.
A Dec-POMDP can be formally described as a tuple ${\cal M}$, i.e.,
\begin{equation}
{\cal M} = \left\langle {{\cal I},{\cal S},{\boldsymbol{\cal A}},{\cal T},{\cal R},{\cal O},\boldsymbol{\Omega} ,\gamma } \right\rangle, 
\end{equation}
where ${\cal I} = [1,2, ..., n]$ is a set of $n$ agents. $\cal S$ is a finite set of states $\boldsymbol{s}$. $\boldsymbol{\cal A} = { \times _i}{{\cal A}_i}$ means joint action space, and $\boldsymbol{\Omega} = { \times _i}{{\Omega}_i}$ indicates the set of joint observations. Note that the symbol '${\times}$' means Cartesian product. At each time step, each agent $i \in \ {\cal I} $ selects an action $a_i \in {\cal A}_i $, forming a joint action $\boldsymbol{a} = [{a_1},{a_2},...,{a_n}]$. Executing $\boldsymbol{a}$ makes the environment transit from state $\boldsymbol {s}$  to $\boldsymbol {s}'$ with probability $P\left( {\boldsymbol {s}'|\boldsymbol {s},\boldsymbol{a}} \right) = {\cal T}(\boldsymbol {s},\boldsymbol{a},\boldsymbol {s}')$, where $\cal T$ is a set of transition functions. Since an agent can partially observe the environment, it receives an individual observation $o_i \in \Omega_i$ at each time step. The environment forms a joint observation that can be represented as $\boldsymbol{o}=[{o_1},{o_2},...,{o_n}]$. The observation function ${\cal O}$ defines a probabilistic mapping from $s’$ and $\boldsymbol{a}$ to the observations $\boldsymbol{o}$ received by all agents, i.e., $P\left( {\boldsymbol{o}|\boldsymbol {s}',\boldsymbol{a}} \right) = {\cal O}(\boldsymbol{o},\boldsymbol {s}',\boldsymbol{a})$. In fully cooperative scenarios, all agents share a collective reward ${r_t} = {\cal R}({s_t},{\boldsymbol{a}_t}) \in \mathbb{R}$ at time step $t$. Each Agent is trained to learn a policy $\mu_i$ that maximizes expected return $\mathbb{E} \left[ \sum\nolimits_{t = 0}^H {{\gamma _t}} {r_t}  \right]$, where $H$ is time horizon, and $\gamma$ means discount factor, with values closer to 0 making the agents more myopic and values closer to 1 making them more far-sighted. 
In a multi-agent system, if all policy agents have the same action space and observation function, i.e.,
${\cal A}_1={\cal A}_2=...={\cal A}_n$ and ${\Omega}_1={\Omega}_2=...={\Omega}_n$, the system is homogeneous, otherwise the system is heterogeneous. 

However, solving Dec-POMDP poses significant challenges for conventional methods and single agent RL algorithms, because Dec-POMDP is proved to be nondeterministic exponential (NEXP) complete \citep{bernstein2002complexity, oliehoek2016concise}, which suggests the time required to find a solution grows exponentially with respect to the joint action and observation spaces in worst case. 
\subsection{MADDPG as a solution to Dec-POMDP}
With the rising of artificial intelligence, MARL has been developed to handle Dec-POMDP. Now we propose to use a policy-based MARL algorithm, MADDPG, as the framework to address multiple ship cooperative tasks. To stabilize learning, we also incorporate learnable communication protocols into MADDPG.

\begin{figure*}
	\centering
		\includegraphics[scale=.32]{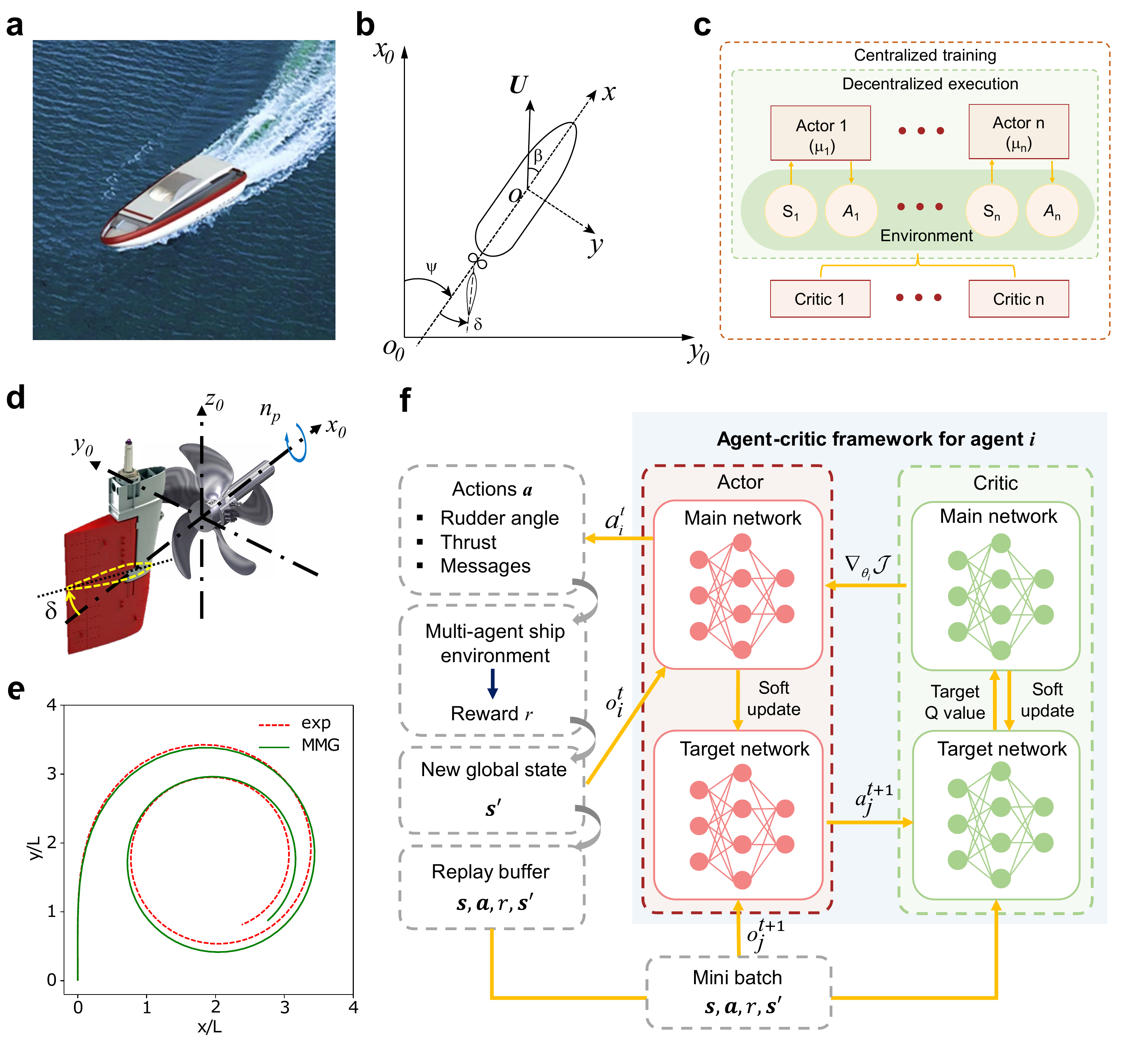}
	\caption{(a) An illustration of USV (b) Reference system for modeling of ship dynamics (c) Overview of centralized training and decentralized execution approach. Each agent $i$ receives observations $o_i$ and executes actions $a_i$. A central critic has access to the local observations and actions of all agents during the training stage. (d) Propeller revolution rate and rugger angle control the ship's movement. (e)The ship trajectory predicted by the MMG model shows good agreement with the experiment in the turning test. (f) MADDPG framework for agent $i$ in our multi-ship environment. Both the actor and the critic contain two subnetworks, namely one main network and one target network. MADDPG allows agents to communicate. When one agent generates a message at time step $t$, this message could be used as part of observation for other agents}
	\label{FIG:2}
\end{figure*}

MADDPG extends deep deterministic policy gradient (DDPG) to multi-agent system. As shown in Figure \ref{FIG:2}(c), MADDPG embraces the paradigm of CTDE \citep{foerster2016learning}. Each agent in MADDPG has a pair of actor and critic networks. During the training phase, the centralized critic network has access to the local observations and actions of all agents, then evaluating the expected return of taking certain actions in a given state. Once the agents have been trained, critics can be discarded. Execution is decentralized, so the actor network maps the agent's local observation to actions. In addition to maneuvering skills, agents may also learn to generate messages that will be received by other agents at next step as part of local observation.  The incorporation of communication within the CTDE framework helps address challenges like non-stationarity \citep{papoudakis2019dealing}, coordination \citep{foerster2016learning, du2021learning}, and scalability \citep{egorov2022scalable}, making it a prevailing approach in MARL.

We illustrate the actor-critic architecture of MADDPG in Figure \ref{FIG:2}(f). Both the actor and critic contain a main network and a target network \citep{lillicrap2015continuous}. Assume that an agent $i$ has a policy called actor $\mu_i$ that is parameterized by $\theta_i$. The main network of actor will produce an action based on given observation, i.e., ${\mu _i}\left( {\cdot|\theta _i^\mu } \right):{\cal O}_i \mapsto {\cal A}_i$ . We add Ornstein-Uhlenbeck noise ${\cal W}_t$ to the produced action to encourage exploration, so at time step $t$ the new action can be written as:
\begin{align}
    a_i^t = {\mu _i}\left( {o_i^t|\theta _i^\mu } \right) + {\cal W}_t.
\end{align}
Then, each agents executes the generated action $a_i$, and receives a collective reward $r_i$ from environment due to the fully cooperation setting. After that, the environment transits from state $\boldsymbol{s}$ to next state $\boldsymbol{s'}$. The tuple composed of $[\boldsymbol{s},\boldsymbol{s'}, \boldsymbol{a}, r_t]$ is restored in replay buffer ${\cal D}$ as an experience. Note that in simplest case, $\boldsymbol{s}$ assembles all agents' observations, $\boldsymbol{s} = \left( {o_1},{o_2},...,{o_n} \right)$. However, state $\boldsymbol{s}$ can include more state information when necessary.  

After accumulating enough experiences in the replay buffer, a batch of experiences is randomly sampled for learning. Let  ${Q _i}\left( {\cdot|\theta _i^Q } \right)$ be the centralized critic network that uses global information, including the actions of all agents $\boldsymbol{a} = [{a_1},{a_2},...,{a_n}]$ and state observation $\boldsymbol{s}$, as input. Then, it outputs the Q-value of agent i via a neural network. Therefore, the target Q-value can be calculated as:
\begin{align}
y_i & = r_i + \gamma Q_i' \left( {\boldsymbol{s}', \boldsymbol{a}'} |\theta _i^{Q'} \right ).
\end{align}
The critic network is updated by minimizing the loss function:
\begin{align}
{\cal L}(\theta _i^Q)={\mathbb{E}_{\boldsymbol{s}',\boldsymbol{s},\boldsymbol{a},r}} \left[{(Q_i (\boldsymbol{s},\boldsymbol{a} |\theta _i^Q) - y_i)^2}\right].
\end{align}
The gradient of the expected return with respect to the actor's parameters is computed by:
\begin{multline}
{\nabla _{{\theta _i}}}{\cal J}\left( \theta _i ^ \mu\right) \\
= {\mathbb{E}_{\boldsymbol{s},\boldsymbol{a}\sim D}}\left[{\nabla _{{\theta _i}}}{\mu _i}\left( {o_i|\theta _i^\mu } \right) \times {\nabla _{{a_i}}}Q_i \left( {\boldsymbol{s},\boldsymbol{a}|\theta _i^Q} \right){|_{a_i = {\mu _i}\left( {o_i|\theta _i^\mu } \right)}}\right] 
\end{multline}
This gradient suggests how to change the actor's parameters to increase the expected return. The update rule is:
\begin{align}
{\theta _i} \leftarrow {\theta _i} + \alpha {\nabla _{{\theta _i}}}{\cal J},
\end{align}
where $\alpha$ is the learning rate. The parameters of a target network for agent $i$ are soft updated from the corresponding main network:
\begin{align}
\theta _i' \leftarrow \tau \theta _i + \left( {1 - \tau } \right)\theta _i'.\   
\end{align}

The above-mentioned process is repeated for many episodes. The replay buffer continues to collect new experiences, and the actor updates its policy based on the feedback from the critic. By iteratively updating the actor network in this way, the policy gradually improves, leading to actions that yield higher returns. The whole process is summarized in algorithm 1.

It is worth to mention that the action space of MADDPG is designed for an environment with continuous action spaces. However, there are some scenarios in which we want MADDPG output discrete actions, e.g., producing digital signal that consists of binary values. In these cases, we employ the Gumbel-Softmax operation \citep{jang2016categorical} to convert the continuous action space to a discrete one. Since the softmax function is differentiable and approximates the discrete distribution in a continuous way, we can still calculate the policy gradients and update actor parameters as usual. 

\end{multicols}

\begin{algorithm*}[!h]
\caption{MADDPG for multiple ships coordination tasks}\label{algorithm}

\textbf{Input:} {Dimensions of actor and critic networks, batch size ${\cal B}$, training episode $M$, maximum steps per episode $L_e$, Ornstein-Uhlenbeck noise ${\cal W}$}, number of agents $n$, replay buffer ${\cal D}$ with size ${\cal C}$.

\textbf{Outputs:} {Policy ${\mu} _i$ for each agent}.

\textbf{Initialize:}{weights of actor and critic network, i.e., $\theta_i^\mu$ and $\theta_i^Q$.}

\For {episode = 1 to ${\cal M}$} 
{Initialize a random process $\cal W$ for action exploration\;
$\boldsymbol{o}$\;
\For {t=1 to $L_e$}
{for each agent $i$, select action $ a_i^t = {\mu _i}\left( {o_i^t|\theta _i^\mu } \right) + {\cal W}_t.$ with respect to current policy and exploration\;
Execute actions $\boldsymbol{a}=\left(a_1, a_2, ..., a_n\right)$, and observe reward $r$ and new observation $\boldsymbol{s}'$\;
Store $\left(\boldsymbol{s}, \boldsymbol{a}, r, \boldsymbol{s}' \right)$in replay buffer ${\cal D}$\;
$\boldsymbol{s}\leftarrow \boldsymbol{s}'$\;
\For {agent $i$=1 to $n$}
{Sample a random minibatch with ${\cal B}$ samples $\left(\boldsymbol{s}, \boldsymbol{s}', r, \boldsymbol{a} \right)$ from ${\cal D}$\;
Set $y_i = r_i + \gamma Q_i'\left( {\boldsymbol{s}', \boldsymbol{a}'} |\theta _i^{Q'} \right)$ \;
Update critic by minimizing loss: ${\cal L}(\theta _i^Q)={\mathbb{E}_{\boldsymbol{s}',\boldsymbol{s},\boldsymbol{a},r}} \left[{(Q_i (\boldsymbol{s},\boldsymbol{a} |\theta _i^Q) - y_i)^2}\right].$ \;
Update actor using the sampled policy gradient: 

${\nabla _{{\theta _i}}}{\cal J}\left( \theta _i ^ \mu\right) = {\mathbb{E}_{\boldsymbol{s},\boldsymbol{a}\sim D}}\left[{\nabla _{{\theta _i}}}{\mu _i}\left( {o_i|\theta _i^\mu } \right) \times {\nabla _{{a_i}}}Q_i \left( {\boldsymbol{s},\boldsymbol{a}|\theta _i^Q} \right){|_{a_i = {\mu _i}\left( {o_i|\theta _i^\mu } \right)}}\right]$ }
Update target network parameters for agent $i$: $\theta _i' \leftarrow \tau \theta _i + \left( {1 - \tau } \right)\theta _i'\ $}
}
\end{algorithm*}

\begin{multicols}{2}
\subsection{Ship dynamics}
Ship dynamics is a fundamental part in building a simulated environment. Figure \ref{FIG:2}(a) illustrates an USV with length $L$, which we utilize in the MARL environment. We list ship particulars in Appendix A.  We employ mathematical model groups (MMG) to simulate ship movements, since it offers a relatively precise way to evaluate the trajectory by considering various kinds of hydrodynamic characteristics \citep{yasukawa2015introduction}. In MMG, the ship maneuvering dynamics are governed by three factors, namely propeller thrust, rudder angle, and hydrodynamic forces acting on the ship. We consider a 3-degree-of-freedom (3-DOF) ship maneuvering problem where the primary ship motions are characterized by surge, sway, and yaw. As a result, the motion equations can be expressed as follows:

\begin{align}
\begin{aligned}
\left( {m + {m_x}} \right)\dot u - \left( {m + {m_y}} \right){v_m}r - {X_G}m{r^2} &= {X_H} + {X_R} + {X_P},\\
\left( {m + {m_y}} \right){{\dot v}_m} - \left( {m + {m_x}} \right)ur + {X_G}m{{\dot r}^2} &= {Y_H} + {Y_R},\\
\left( {{I_{zG}} + X_G^2m + {J_z}} \right)\dot r + {X_G}m\left( {{v_m} + ur} \right) &= {N_H} + {N_R}, \nonumber
\end{aligned}
\end{align}
where \(X_H, Y_H, N_H\) represent the hydrodynamic surge force, lateral force, and yaw moment acting on the hull of the ship, respectively. \(X_R, Y_R, N_R\) represent the surge force, the lateral force, and the yaw moment in midship by steering. The terms \(m_1, m_x, m_y\) denote the mass of the ship and the added mass in the direction of the x-axis and the y-axis. The velocities \(u, v, r_m\) correspond to the surge, lateral, and rotational speeds at the center of the lateral resistance of the ship. The variable \(x_G\) specifies the longitudinal position of the center of gravity of the ship. The moments of inertia about the center of gravity and along the vertical axis are given by \(I_{zG}\) and \(J_z\), respectively.

As shown in Figure \ref{FIG:2}(b), we establish a coordinate system to describe the movement of USVs. $o_0-x_0y_0z_0$ defines fixed earth reference frame, where the $x_0y_0$ plane indicates the surface of the still water. The local reference system $o-xyz$ is fixed in the moving ship, where $o$ is the center of the ship, and the axes d x, y and z point to the bow of the ship, starboard and vertically downwards, respectively. We define the angle between the $x_0$ and $x$ axes as the heading angle ${\psi}$. $\beta$ represents drift angle. The motion of the ship is governed by the angle of inclination of the rudder $\delta$ and the revolution rate of the propeller $n_p$, where the rudder angle controls the direction and the propeller regulates the speed (see Figure \ref{FIG:2}(d)). As shown in Figure \ref{FIG:2}(e), we compare the simulated turning circle with the experiment. The turning circle test is conducted with full engine power and the rudder angle is set to 35 °. The results suggest that the MMG model can accurately predict the trajectory of the ship.

\subsection{COLREGs compliance}
International Regulations for Preventing Collisions at Sea (COLREGs) are established by the International Maritime Organization and provide guidelines for ships and other vessels to avoid collisions on the water. COLGRGs defines three situations in which vessels may encounter each other, namely head-on, overtaking, and crossing, as shown in Figure \ref{FIG:4}(b). In addition, COLREGs gives two primary designations to vessels in a potential collision situation: the stand-on vessel and the give-way vessel. 
\begin{itemize}
\item In a head-on situation (Rule 14), both vessels are considered to be give-way vessels and both should alter their course to starboard. 
\item In an overtaking situation (Rule 13), the overtaking vessel (the one coming from behind) is the give-way vessel, while the vessel being overtaken is the stand-on vessel.
\item In a crossing situation (Rule 15), the vessel that has the other on its starboard (right) side is the give-way vessel, while another one is stand-on vessel. 
\end{itemize}
The give-way vessel does not have the right of way in a given situation, so it should take early and substantial action, e.g., altering its course to starboard to prevent possible collision. The stand-on vessel is expected to maintain its course and speed. However, COLREGs also makes clear that, if essential for collision prevention, the stand-on vessel is permitted to adjust its direction or speed as an evasive measure (Rule 17).

\subsection{Maritime communication}
\label{maritime comm}
Reliable communication among vessels, shore stations, and other maritime assets is essential for navigation safety and operational efficiency. However, noise can be inevitable in maritime communication. To investigate how a noisy channel can affect the learning process, we introduce additive white Gaussian noise (AWGN) and binary symmetric channel (BSC).

BSC deals with discrete signal. Consider a message sent by agent $i$ has $k$ signals, $\boldsymbol{m}_i = \left[ m_1, m_2,...m_k \right]$, in BSC all transmitted signals can only take binary values, i.e., for the $j$-th transmitted signal, we have ${m}_j \in \{0,1\}$. BSC is characterized mainly by the probability of a bit flip error $P_{e}$. It is the probability that any given bit will be received in error, that is, a '0' is flipped to a ‘1’ or a ‘1’ is flipped to a ‘0’, during transmission. Therefore, the signal received from the channel $j$ can be written as ${\hat m}_j = {m_j} \oplus {n_e}$, where $n_e$ is determined by Bernoulli($p_e$) and $\oplus$ denotes XOR operation.

AWGN deals with continuous signals, so for $j$-th signal, the transmitted bit $m_j$ should be the real value number, which means $m_j \in \mathbb{R}$. Then, a white noise, which follows a Gaussian distribution ${\cal N}(0, \sigma^2)$, will be added to the transmitted signal. Therefore, after passing through the noisy communication channel, the output of the $j$-th channel is ${\hat m_j} = {m_j} + {\cal N}(0, \sigma^2)$.

To quantify the strength of a signal relative to the background noise, we introduce a non-dimensional parameter called signal-to-noise ratio (SNR). In AWGN channel, the SNR is defined as:
\begin{align}
 {SNR} = \frac{P_{\text{signal}}}{P_{\text{noise}}},
\end{align}
where \( P_{\text{signal}} \) denotes the power of the signal and \( P_{\text{noise}} \) represents the power of the background noise. Higher SNR suggests that the signal is much stronger than the noise, leading to better clarity. SNR is often expressed in decibels (dB) for convenience. The formula to convert SNR to decibels is:
\begin{align}
{{SNR}\left(dB\right)} = 10 \log_{10} \left( \frac{P_{\text{signal}}}{P_{\text{noise}}} \right).
\label{eq:SNR-dB}
\end{align}
If we normalize the input message  $\boldsymbol{m}_i = \left[ m_1, m_2,...m_k \right]$ to an average power of 1, we have
\begin{align}
m_j^{normalized} = \sqrt{k} \frac{m_j}{\sqrt{{\mathbf{m}_i^\top}{\mathbf{m}_i} }}, j = 1,2,...,k.
\end{align}
The SNR in equation.\ref{eq:SNR-dB} can be simplified to:
\begin{align}
SNR(dB) =-10\log_{10}{\sigma ^2}.  
\end{align}

\section{Environment Setup}
To examine the performance of our proposed MARL method in solving cooperative tasks of multiple ships, we fabricate two virtual scenarios: cooperative navigation and cooperative collision avoidance. Due to the restriction on observability, communication protocols must be established between agents. We develop our code based on the well-established Multi-Agent Particle Environment by OpenAI, which is simple but widely considered a standard platform for testing MARL algorithms. 
\subsection{Ship cooperative navigation}
In maritime settings, a fundamental scenario involves human operators sending instructions to USV, directing them to perform specific tasks \citep{wang2021cloud}, e.g., setting waypoints and reaching a goal. Such communication messages are often formatted in a predefined way to ensure a proper interpretation \citep{bostrom2020mind,amro2023communication}. However, to improve collaboration, efficiency, privacy,  and reduce the need for human interactions, we sometimes hope that USV can learn communication protocols through training. 

Hence, we create a scenario called ship cooperative navigation to simulate the process in which a human proxy sends instructions to ships. As illustrated in Figure \ref{FIG:2}(a),  this scenario contains $M$ landmarks and $N$ ships and a speaker (proxy). For landmarks, they are randomly placed in a 2D plane, and each of them has a distinctive color. Each ship can observe its own status and relative positions to all landmarks, but it does not know which target it should navigate to. In contrast, the speaker knows the correct landmark color, so the observation of speaker $\boldsymbol{o}_s$ is a one-dimensional vector that composed all target colors. For example, when $N=1$ and target color is pure green, we have $\boldsymbol{o}_s = \left[0,1,0\right]$. The speaker must learn to send explicit messages with $k$ elements $\boldsymbol{m} = \left[ m_1, m_2,...m_k \right]$, which can be regarded as embeddings of raw observations $\boldsymbol{o}_s$, to the ships in order to inform them of the correct target. For example, we set $k=2$ when $N=1$ and $M=3$. Then, other ships receive messages $\boldsymbol{\hat{m}}$ from the speaker, together with self-status and relative positions to all landmarks to form ship's observation $\boldsymbol{o}_u = \left [U_x, U_y,\beta, \delta, \theta_1, \theta_2,..., \theta_m, d_1, d_2,..., d_m,  \boldsymbol{\hat{m}} \right]$, where $U_x$ and $U_y$ respectively means velocities of x and y direction in O-xyz system, $\beta$ indicates drift angle, $\delta$ is current rudder angle, $d_1, d_2, ..., d_n$ and $\theta_1, \theta_2, ..., \theta_n$ measure relative distances and angles from own ship to all landmarks. Then ships use individual observation $\boldsymbol{o}_s$ as input to generate actions. The action space contains propeller revolutions $n_p$ and desired rudder angle $\delta _d \in [-35^\circ, 35^\circ]$. It may take several time steps for the rudder to reach such desired angle. Since the agents have different action space and observation space, this system is heterogeneous. Ships and speaker will receive a same distance related reward $r_{goal}$ at each time step. The total reward of all agents will be sent to the critic network. $r_{goal}$ is defined as:
\begin{align}
r_{\text {goal }}=\left\{\begin{array}{cc}
r_{\text {goal }} & \text { if }\left\|\boldsymbol{P}_t-\boldsymbol{P}_{\text {goal }}\right\|<10 \\
\lambda_{\text {goal }}\left\|\boldsymbol{P}_{t}-\boldsymbol{P}_{\text {goal }}\right\| & \text { otherwise }
\end{array},\right. 
\label{rgoal}
\end{align}
where $P_t$ means current ship position and $P_{goal}$ indicates goal position.
We summarize the network structures and hyper parameters used in Table \ref{Table:1}.
\end{multicols}    

\begin{table}[h!]
\centering
\caption{Hyperparameters for Cooperative navigation and Cooperative collision avoidance.}
\begin{tabular*}{\tblwidth}{@{} LLLL@{} }
\toprule
\textbf{Hyperparameter} & \textbf{Cooperative navigation} & \textbf{Cooperative collision avoidance} \\ 
\midrule
Neurons of Hidden layers for actor network & (64, 64, 64) & (256, 256) \\ 
Neurons of Hidden layers for critic network & (64, 64) & (128, 128) \\ 
Learning rate of actor, critic & 0.002, 0.01 & 0.0005, 0.001 \\ 
Optimizer & Adam & Adam \\ 
Activation function & Relu & Relu \\ 
Soft update rate & 0.001 & 0.001 \\ 
Replay buffer size & 1.00E+06 & 1.00E+06 \\ 
Batch size & 1024 & 512 \\ 
Discount factor & 0.95 & 0.95 \\ 
Maximum steps per episode & 25 & 40\\
\bottomrule
\label{Table:1}
\end{tabular*}
\end{table}

\begin{multicols}{2}
\subsection{Ship cooperative collision avoidance}
As shown in \ref{FIG:4}(a), in this scenario, two ships must collaborate to prevent collisions according to COLREGS. We assume that ships are invisible to each other, so they have to communicate to alleviate partial observation. For each ship, the continuous action space includes three elements, the propeller revolution $n_p$, the desired rudder angle $\delta_d$, and the message $\boldsymbol{m}$, i.e., $\boldsymbol{a}=[n_p, \delta_d, \boldsymbol{m}]$. The message $\boldsymbol{m}$ is a mapping of individual observations, and will be sent to the other ship at next time step. In order to examine whether the agent can extract or compress key information from observation, we limit communication bandwidth to three bits $\boldsymbol{m} = {[m_1,m_2, m_3]}$. Current ship position $\boldsymbol{P}_t = [p_x, p_y]$, rudder angle $\delta$, bow angle $\Psi$, speed $[U_x, U_y]$, drift angle $\beta$, relative positions to its designation $[d, \theta]$, and messages received by the other ship at next time step $\bold{\hat{m}}$ constitute the observation space $\boldsymbol{o} = [\delta, \beta, \Psi, p_x, p_y, d, \theta,\bold{\hat{m}}]$. We suppose agents can get perfect observation of self status, e.g., current position $\boldsymbol{P}_t$, but communicated message will be perturbed by an AWGN type noisy channel during transmission. Since both ships have identical action and observation space, this multi-agent system is homogeneous. To obtain the optimal reward, the two ships must coordinate to reach goal without collision. As we mentioned in section 3.1, the total reward for the system $R_c$ is defined as the sum of ship A's individual reward $r_a$ and ship B's reward $r_b$ to facilitate cooperation, i.e.,
\begin{align}
 {R_c} = r_a + r_b.
\end{align}
For each ship, the reward function $r$ consists of three parts: namely goal reaching reward, COLREGs reward and collision reward. We have:
\begin{align}
r = r_{goal} + r_{COLREGs} + r_{collision},
\end{align}
where the goal reaching reward $r_{goal}$ directs the ship's navigation towards its objective. It has the same form as equation.\ref{rgoal}.

$r_{collision}$ can be expressed as:
\begin{align}
r_{\text {collision }}=\left\{\begin{array}{cc}
-r_{\text {collision }} & \text { if }\left\|\boldsymbol{P}_t-\boldsymbol{P}_{\mathrm{ts}}\right\|< 2S_r \\
0 & \text { otherwise }
\end{array}.\right.
\end{align}
We use $S_r$ to indicate radius of ship domain where other ships are not supposed to enter. Both ships will receive a same penalty if the distance between two ships is less than $2S_r$. To simplify model, we set $2S_r = L$. $\boldsymbol{P}_{ts}$ means the position of another ship. 

In addition, both ships can receive a reward if two ships avoid collisions in a COLREGs complaint way. We have:

\begin{align}
r_{\text {COLREGs }}=\left\{\begin{array}{cc}
r_{\text {COLREGs }}, & \text { if give way ship turns right }, \\
-r_{\text {COLREGs }}, & \text { otherwise }.
\end{array}\right.
\end{align}

As shown in Figure \ref{FIG:4}(b), we define the intersection of trajectories as point $P$, if the two ships continue on their rudder angles and speed without change. Collision may happen at this point. Both ships are randomly placed between two circles with radii $R1$ and $R2$ before training start, resulting in three encounter scenarios defined by COLREGS, namely head on, crossing, and overtaking. To train a robust strategy, we manually set the occurrence probabilities of all encounter scenarios to be equal. See Table \ref{Table:1}
for network parameters we use in this scenario.
\section{Results and discussion}
To evaluate the effectiveness of our proposed method, we apply it to the tasks of ship cooperative navigation and cooperative collision avoidance. We compare the training outcomes against benchmark methods. Several communication-related challenges, including the impact of a noisy channel, deciding when to communicate, and what information should be exchanged, are also discussed.

\begin{figure*}[H]
	\centering
	\includegraphics[scale=.255]{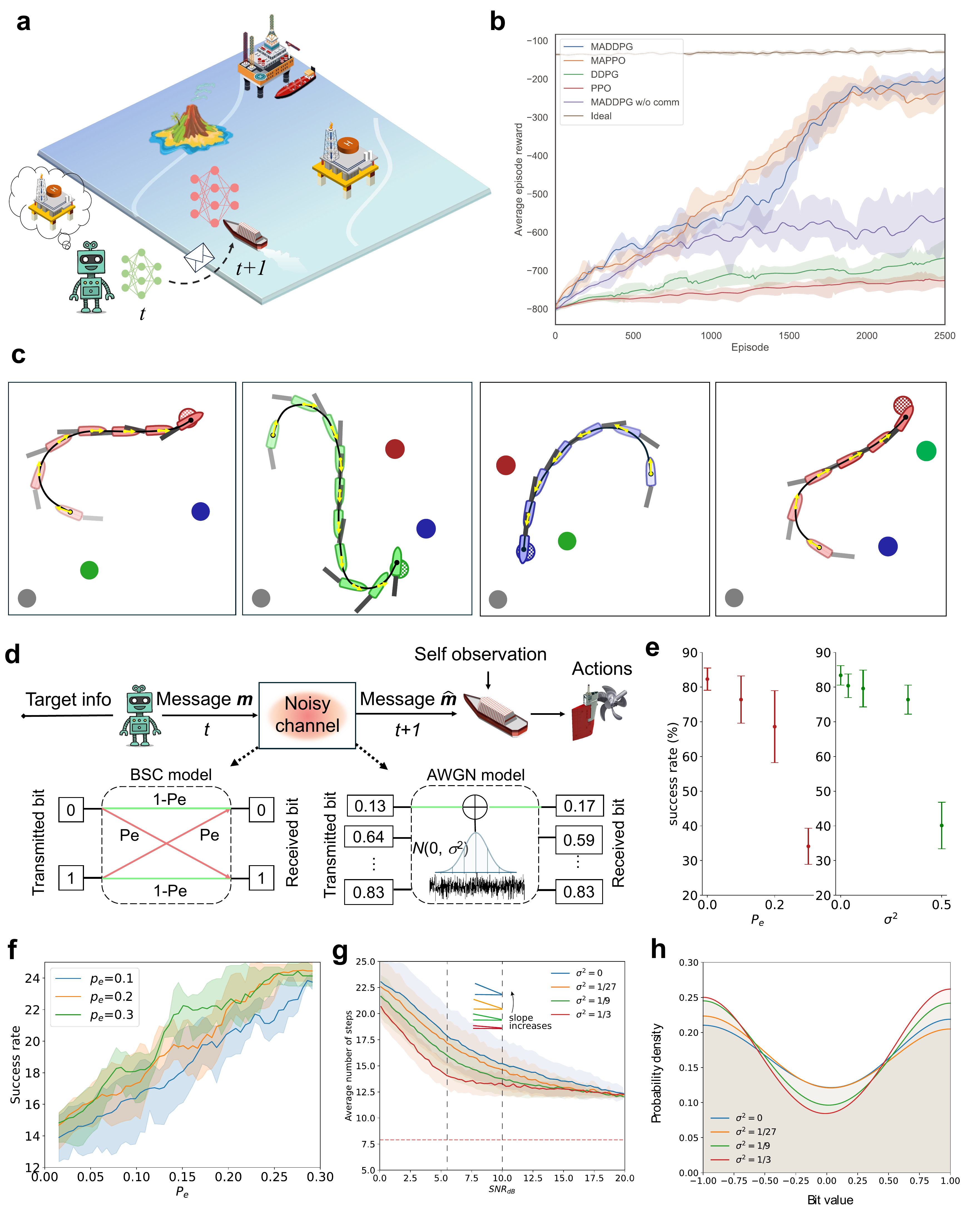}
	\caption{In ship cooperative navigation scenario (a), ships can observe relative position to all landmarks, but it doesn’t know which one is real target. The human proxy (speaker) knows the color of the real target, so it must learn to output messages that guide ships to reach corresponding targets.(b) The reward history of MADDPG against other methods. (c) Typical results for $N$ = 1 case. The gray circle represents human proxy (speaker), and colored circles are landmarks. Please refer to movie S1 and S2 in the supplementary material for animations of additional cases. (d) An illustration of noisy channels. In BSC channel, we impose flip error on discrete digital signal, while in AWGN channel, the Gaussian white noise is added. (e) The training success rate decreases with $P_e$ and $\sigma^2$, in BSC (left) and AWGN (right) channel respectively  (f) Average number of steps required to reach the goal increases with $P_e$ in the BSC channel (green solid line) for $N=1$ case. The red dot line means minimum steps that we estimate through the PID controller.(g) The effects of additive noise on average number of steps. }
	\label{FIG:3}
\end{figure*}
\subsection{Ship Cooperative Navigation}

We evaluate the efficacy of the MADDPG algorithm in a cooperative navigation scenario when $N$=1. Our analysis benchmarks MADDPG against four distinct methods: 1) a PID controller, 2) MAPPO, 3) a conventional DDPG algorithm, and 4) MADDPG without communication. Specifically, the details are as follows:
\begin{itemize}
    \item \textbf{PID Controller}: Initially, we implement a PID controller to estimate the optimal reward. In this setup, ships are presumed to be aware of their actual objectives. We define the relative error, \( e(t) \), as the angular discrepancy between the target direction and the ship's heading angle. Here, the ship's propeller rotates at its maximum frequency, and the objective is to minimize this relative error through rudder angle adjustments. The controller's output is formulated as 
    \[ u(t) = K_p e(t) + K_i \int_0^t e(\tau) \, d\tau + K_d \frac{de(t)}{dt}, \]
    where \( K_p \), \( K_i \), and \( K_d \) represent the coefficients for the proportional, integral, and derivative terms, respectively, fine-tuned through a trial-and-error process.

    \item \textbf{MAPPO}: We then examine the MAPPO \citep{yu2022surprising}, a policy-based MARL algorithm. MAPPO extends the principles of its single-agent counterpart, PPO, which is prevalently used in automatic collision avoidance and ship path planning. Contrary to MADDPG, MAPPO operates on an on-policy basis, implying that the data for policy updates is consistently sourced from the latest policy iteration.

    \item \textbf{DDPG}: The study also incorporates the conventional DDPG algorithm. Here, each agent is equipped with an individual DDPG network. Owing to the absence of a centralized critic, agents undergo training in a distributed manner, emphasizing the algorithm's decentralized nature.

    \item \textbf{MADDPG with Blocked Communication}: Lastly, to underline the significance of communication in this context, we assess the performance of MADDPG under conditions where communication channels are completely obstructed. This approach aims to illustrate the impact of communication barriers on the effectiveness of cooperative strategies in complex scenarios.
\end{itemize}

As shown in Figure~\ref{FIG:3}(b), MARL algorithms, such as MADDPG and MAPPO, demonstrate superior performance compared to other methods by learning to achieve higher rewards, approaching the ideal reward we obtain from the PID controller after approximately 2,500 episodes of training. The ship controlled by MADDPG policy is able to reach the real landmark around 84.2\% of the time. However, MADDPG's effectiveness significantly diminishes when communication between the human proxy and USVs is disrupted. Although the reward metrics show an improvement from a randomly initialized strategy after training, USVs mainly exhibit unproductive behaviors such as spinning in place rather than moving towards the target when communication is blocked. The overall success rates are less than 5\%. In addition, traditional single-agent RL algorithm, DDPG, also fails to learn correct behaviour,this problem arises from issues with credit assignment and the inherently non-stationary nature of the environment. 

Next, we investigate the maneuvering skills acquired by USVs. Figure \ref{FIG:3}(c) shows four typical results of ship cooperative navigation scenario. The black bar at the tail of ship represents rudder angles, and the yellow arrows indicate relative velocity at each step. We find USVs initially engage in significant rudder adjustments to align their bow direction with the target as accurately as possible. Subsequently, they perform counter-steering maneuvers to eliminate lateral velocity, thus enabling straight-line travel. As a USV nears its target, it is noted that there is a reduction in the propeller's thrust, which aids in deceleration. Concurrently, the rudder is deflected towards the target side to facilitate goal attainment. This strategy is considered effective and rational, especially when we consider navigation dynamics. Specifically, at higher speeds, a ship's turning radius is expanded due to increased inertia. By reducing speed, USVs can execute more precise turns, thereby enhancing their probability of successfully reaching the target.

\subsection{Noise Limitations}
We explore the impact of noise on communication in USVs, particularly focusing on how unpredictable noise interference affects message transmission. BSC method and AWGN model are utilized to simulate natural random processes for discrete signals and continuous signals, respectively (See Figure \ref{FIG:3}(d)).

We first train policy networks under four different error probabilities under four different error probabilities. As shown in Figure \ref{FIG:3}(e), the success rate decreases with $P_e$. As $P_e$ increases, the likelihood of each transmitted bit being flipped also increases, which means more bits are likely to be received incorrectly. Once ships receive wrong message, it may navigate to wrong destination, thus reducing success rate. When $P_e =0.3$,the communication channel becomes so noisy that ship can not receive meaningful information, causing the listener fails to learn a reliable policy. Then, we test the trained policies in across a range of $P_e$ to access their robustness. Results are plotted in Figure \ref{FIG:3}(f). We notice that for all trained policies, the steps needed to fulfill task generally increases with $P_e$ before reaching maximum steps per episode $L_e$=25. This finding is accordance with Figure  \ref{FIG:3}(e), because ships basically require more trails to reach goal. Failures happen when ships do not reach the given target within $L_e$ steps. Besides, the slopes of the four curves are roughly consistent, which suggests even if noise is considered during training, the trained policy cannot effectively resist the flip errors in BSC channel.   

For AWGN channel, we also train policies under different noise level, then testing these trained policies across a range of SNR. We plot the average number needed to reach goal as a function of SNR for different policies in Figure \ref{FIG:3}(g). When SNR = 20, all policies converge to a value around 12.5, which means the impact of negligible at this level. As we expect, a ship requires more steps to reach goal as SNR decreases. This reflects the fact that the noise deteriorates the quality of communication. However, we notice that compared with $\sigma$=0 case, the slope of steps with respect to SNR for $\sigma$=1/3 case is lower. So, strategy trained in higher noise level becomes less vulnerable to additive noise. To some extent, the neural network can learn to adapt to noise, resulting in a more robust strategy. When SNR lowers than 5.8, the channel becomes so noisy that all strategies fail to communicate useful information. Therefore, the steps required to reach the goal show linear growth. 

To investigate why policies trained in higher noise levels can better adapt to noisy channels, we train the network using 12 different sets of random seeds for each noise level and record the output messages throughout the entire training process. Figure~\ref{FIG:3}(h) illustrates the probability distribution of message values generated under various noise conditions. It is evident that as the noise level increases, the speaker tends to produce information with more distinct boundary values, such as \([1, -1]\) to represent a blue landmark, rather than more nuanced values like \([-0.1, 0.2]\). This approach effectively amplifies the differences between the information, thereby mitigating the impact of channel noise.

\subsection{Ship cooperative collision avoidance}

\begin{figure*}
	\centering
	\includegraphics[scale=.27]{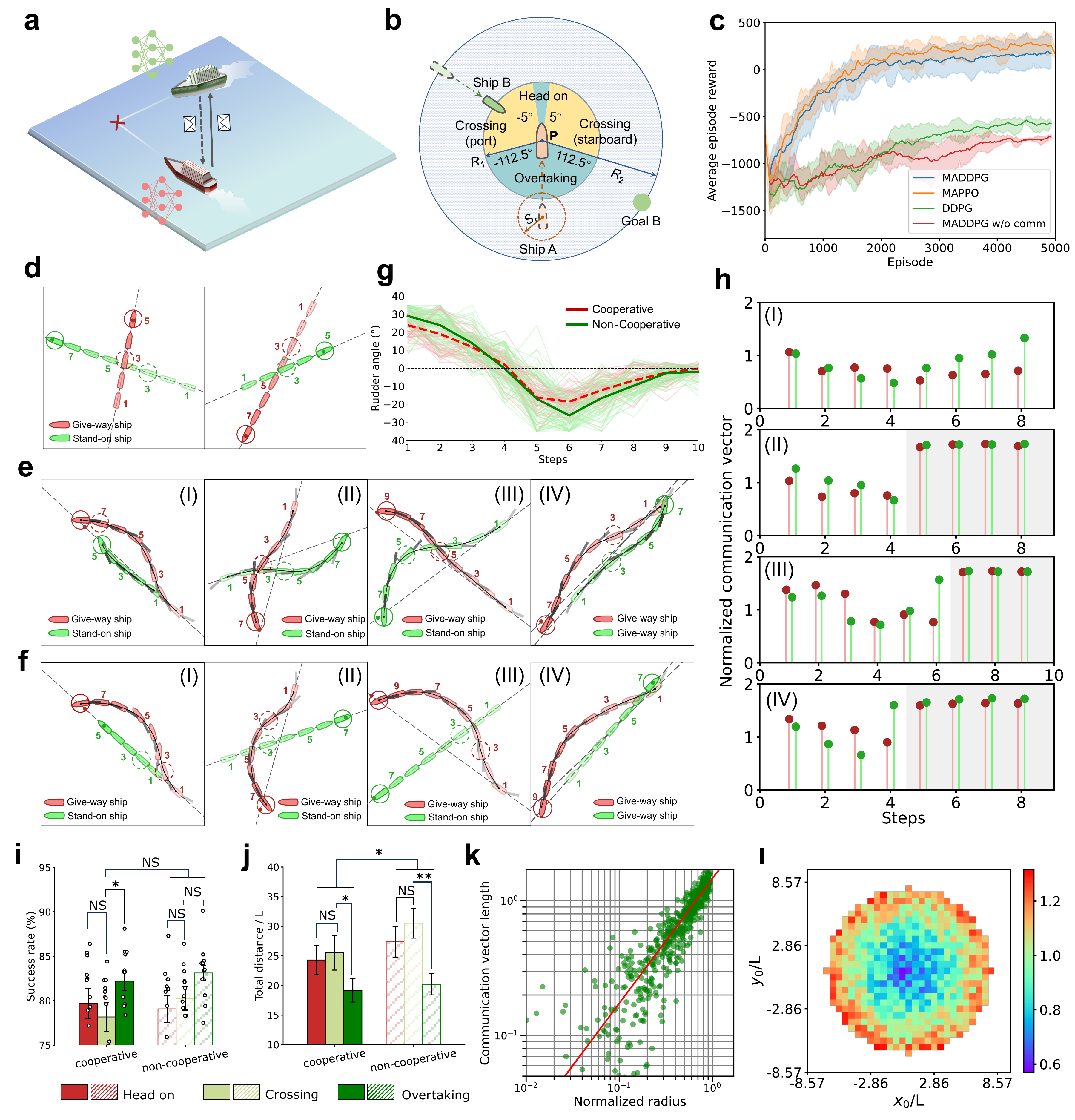}
	\caption{In cooperative collision avoidance scenario (a), both ships are required to reach the goal and aviod collision in a COLREGs complaint way. However, they are constrained by partial observability, which means can only observe own status and relative positions to target. Apart from communication, one ship cannot obtain any status information about another ship. (b) Encounter situations defined by COLREGs. (c) The reward history of MADDPG against other methods.(d) Bow crossing may happen when one ship is close to meeting point while another is far away. (e) In cooperative collision avoidance mode, both head on ship and give way ship take actions to prevent collision for (I) Overtaking, (II) Port crossing, (III) Starboard crossing, and (IV) Head on scenarios.  (f) In non-cooperative manner, only give way ship alters, rudder angle. The stand-on ship doesn't take measures. (g) Compared with cooperative mode, the give way ship adopts larger rudder angles. (i) T tests suggests that there is no significance in the success rate of cooperative and non-cooperative mode. (j) T test suggests that the total traveling distance will be significant lower if two ships cooperate to avoid collision. (h) Communication vector length of four encounter situations where we illustrate in (e). Ships may adopt different communication patterns before and after two ships meet. (k) Before two ships meet, the communication vector length is highly related to the ship's distance to meeting point. (I) The further distance, the larger the communication vector length. }
	\label{FIG:4}
\end{figure*}

We compare the performance of MADDPG on ship cooperative collision avoidance scenario with three baseline methods as we describe in section 4.1. Figure \ref{FIG:4} (c) shows average returns of different algorithms on ship cooperative collision avoidance task. The solid line represents mean reward over 5000 episodes, while the shaded area indicates one standard deviation over six random seeds. We also use a sliding window with a length of 100 episodes to smooth these curves. As we expect, MADDPG and MAPPO reach upper bound earlier and outperform other baselines. Repeated tests suggest that an agent trained with MADDPG has an 79.3\% probability of reaching the destination without colliding. By contrast, the success rate of DDPG agent is 52.7\%. The reasons are obvious. In this task, the information exchange is crucial, because the blind ships cannot directly observe the states of other agents. Introducing communication can help reduce the uncertainty inherent in the partial observability of the environment. Also, agents can not learn a good policy when communication is abandoned, even if the centralized critic can deal with the non-stationarity and credit assignment issues existing in DDPG, as one ship does not have any clue about another ship's status.

In Figure \ref{FIG:4}(d-e), we present the trajectories of ship generated by the well-trained strategies in different encounter situations. we depict buffer zones with dot circle at the moment when the two ships were closest to each other, and using solid circle to indicate destination. Result suggests that our trained policies successfully guide ships to reach targets without collision. As we can see in Figure \ref{FIG:4}(i), the success rates of cooperative collision avoidance in head-on crossing and overtaking situations are $79.8\%$, $77.3\%$ and $83.4\%$, respectively. 

As shown in Figure \ref{FIG:4}(d). We find that if one ship is close to the anticipated meeting point $P$ while another is far away, both vessels tend to move forward at full speed without changing the rudder angles. As a result, one ship crosses ahead of the other. This behavior is often referred to as bow crossing. Assume $R_a$ and $R_b$ are distance from the meeting point $P$ to the initial position of ship A and B. Repeated tests suggest that ships are likely to adopt this manner when $\left| R_a - R_b \right|> 23.7 m$.  A plausible interpretation is that the policy agents are aware of that collisions are less likely to happen as long as the two ships maintain their directions and move at same speed, because when one ship arrives meeting point, the distance between two ships are still within a safe regime. Keeping moving forward at maximum speed enables them to reduce distance penalties. In addition, given that two ships are invisible to each other, we can infer that communication protocols are established, so they can take specific actions for distinct situations.  

However, at most of the time, ships must alter their course to prevent collisions and adhere to COLREGs, because collisions will become inevitable if ships maneuver at the same velocity. In Figure \ref{FIG:4}(e), we present representative trajectories in four typical encounter situations, namely crossing from port, crossing from starboard, head on, and overtaking.  

To our surprise, in all scenarios, the stand-on ship proactively takes evasive measures and closely collaborates with the give-way ship to avoid a collision, even when moving in a straight line is the most direct route to its destination. Since both the give-way ship and the head-on ship act to prevent a collision simultaneously, we term this approach ``cooperative collision avoidance", as opposed to a ``non-cooperative" mode where the head-on ship continues forward without adjusting its course. According to Rule 17 of the COLREGs, if the stand-on ship finds itself dangerously close to the give-way vessel, it must take actions that best avoid a collision. Consequently, our strategy is in compliance with COLREGs. Additionally, we observe that our learned strategies do not depend on changing speed to prevent collisions. As noted by \cite{shen2019automatic}, altering speed is not an efficient method for collision avoidance because ships require considerably more time to accelerate or decelerate compared to road vehicles. This characteristic is reflected in our MMG model, which results in the learned policy favoring course adjustments over speed modifications for collision prevention.

However, why the MARL agents learn to avoid collisions in a cooperative way $?$ Will it be better than give-way ship keeps out of the way $?$ To answer these questions, we modify the action space of stand-on ship. The rudder angle of stand-on ship is fixed to zero and propeller revolution is set to be $n_p = 100$ rpm, so it can only move in a straight line. The stand-on ship is still able to inform the give-way ship its status via messages, which means the action space of stand-on ship becomes $\boldsymbol{a}=[\boldsymbol{m}]$. Note that other settings, including observation space, and reward function remain unchanged. 

Comparing \ref{FIG:4}(e) with \ref{FIG:4}(f), it's obvious that give-way vessel uses larger degree of steering to avoid collision. For instance, in the overtaking scenario, give-way vessel first steers rudder to rightmost to avoid the stand-on vessel, and once it has fully cleared the stand-on vessel, it will then steer hard left to head towards the destination, thereby generating a snake like trajectory. Repeated tests also support this observation. We perform 100 tests with randomized initial positions using cooperative and non-cooperative modes. We depict the  rudder angle of give-way ship as a function of steps in Figure \ref{FIG:4} (g). It suggests the give-ship tend to adopt lager rudder angle to avoid collision in non-cooperative mode.

We calculate the success rates of task completion under cooperative and non-cooperative modes and conduct a T-test on the results. It helps assess whether any observed differences are statistically meaningful or could have occurred by random chance. In Figure \ref{FIG:4}(i), We find that, compared to crossing and head-on approaches, both strategies had a higher success rate in overtaking. Then, when comparing the overall success rates of the two strategies, we obtained a p-value greater than 0.05, which suggests no statistically significant difference between the two strategies. Next, we compared the total trajectory lengths of the two ships under each strategy, see Figure \ref{FIG:4}(j). We notice that in cooperative mode, the two ships travel shorter distances from a statistical point of view, and there is a confidence level $95\%$ that indicates that this difference is significant. Therefore, we can conclude that because the cooperative mode results in shorter travel distances and thus incurs smaller distance penalties, the agent preferred to adopt the cooperative obstacle avoidance mode.

\subsection{What and when to communicate}
To analyze the communication strategies generated during collision avoidance, we extracted the communication vectors \( m \) from the four scenarios depicted in Figure \ref{FIG:4}(e). Since different initialization seeds might lead to different communication strategies formed by the agents, and following the suggestion by \cite{mao2020learning}, we primarily analyzed the magnitude of the communication vectors. Figure \ref{FIG:4}(h) shows the relationship between the magnitude of the communication vectors and the number of steps from departure to arrival. We observed that as the two ships' positions continuously approached each other, the magnitude of the communication vectors decreased until the point where they were at their closest distance. Therefore, we hypothesize that the magnitude of this communication vector may be positively correlated with the distance of the ships from the potential intersection point \( P \). To test this hypothesis, we performed a linear regression analysis with the distance to point \( P \) as the independent variable and the magnitude of the communication vector before the intersection as the dependent variable, as illustrated in Figure \ref{FIG:4}(k). The results showed that the regression coefficient is greater than 0 (see the red line), indicating a positive correlation between the two variables. Further, we visualize the average norm of the communication vectors over a 31*31 grid in \ref{FIG:4}(l). Note that we only consider the communication vectors before they come to the closet distance point. In places far from the meeting point, the value is large, while it is small near the center. This evidence clearly indicates that the communication vector contains information about its own location.  We consider this result to be reasonable because, in this task, the two ships cannot directly obtain each other's information without communication. To successfully complete the collision avoidance task, it is essential for each ship to share its own position with the other.

However, the communication pattern changes when two ships pass the meeting point. To our surprise, they tend to generate an communication vector where each element approaches the boundary value e.g., [-1,1,1] or [1,1,-1], etc., so communication vector length is close to 1.732. This kind of message no longer changes with ship position. \cite{sukhbaatar2016learning} reported similar patterns. They believe agent prefers not to communicate unnecessary information to other agents, so it generates repeated messages to indicate other agents that such information less important.  We agree with this point of view for two reasons. First, after successfully completing the collision avoidance, the agent needs to reach the target as quickly as possible to obtain the maximum benefit. In other word, information about other agents is no longer important, so agents indicate others that communicated message can be ignored by sending specific signals. Second, one ship can determine whether it has already avoided another ship based on its own observation and communicated message. For instance, as shown in Figure, in step 5, the give-way ship received vectorized message that indicates stand-on ship’s position at step 4. By comparing relative positions to target and stand-on ship, the give-way ship realizes that it has passed astern of another one, which means collisions are less likely to occur at this moment. So, it generates a signal like [-1,1,1] to tell its partner, “Hi dude, we won’t collide with each other. please focus on goal reaching task”. The stand-on ship received the signal at step 6, it understands the meaning of this message, and also produces a message like this.

\section{Conclusion}

We model the working environment of USVs as a partially observable multi-agent system, where entities collaborate to achieve common goals. To address the challenges of credit assignment and non-stationarity, we employ a multi-agent reinforcement learning (MARL) algorithm, such as MADDPG, to train the policy agents. In addition to taking physical actions that directly affect the environment, such as adjusting rudder angles and speed, each agent must also communicate explicitly with its counterparts to mitigate partial observability. To evaluate the performance of our proposed MARL method, we create two scenarios—cooperative navigation and cooperative collision avoidance—using OpenAI's Multi-Agent Particle Environment

In the cooperative navigation scenario, we show that MADDPG outperforms both single-agent reinforcement learning and communication-disabled algorithms. When noise is present in the communication channel, the agents can adapt to additive white Gaussian noise to some extent. In the cooperative collision avoidance scenario, we demonstrate that the ships can successfully establish communication protocols and avoid collisions in a manner compliant with COLREGs. Rather than merely maintaining speed and rudder angles, we find that the head-on ship will take evasive actions to prevent a collision. Statistical evidence suggests that the total traveling distances of the two ships are reduced by 13\% in cooperative mode compared to the non-cooperative mode. Additionally, the communication patterns change before and after the ships meet at sea.

Although we get some surprising results, we have to admit this work is only an initial attempt by applying RL to the field of multiple ship maneuvering. There are many ways to improve or extend this work.

First, we only consider multiple ship cooperation tasks. In fact, in real world, there are not only cooperations, but also competitions. For example, there is a competitive relationship between smuggling boats and patrol ships. Can we design autonomous navigation systems that allow USVs to coordinate internally to capture smuggling boats, even when smuggling boats adopt various evasion strategies?

Second, we use an encoding of current local observations as messages to communicate, in order to deal with partial observation. However, agents can also leverage historical experience or future plans to further facilitate coordination. Recently, the development of large language models (LLM, e.g., ChatGPT) has made remarkable progress. This may bring a new paradigm for ship control problem. The LLM captain can determine collision avoidance plans through negotiation based on their hydrodynamic characteristics, tasks, and level of intelligence. For example, smart ships may take actions to avoid collisions with manned ships; small ships give way to large container ship, rather than rigidly follow COLREGs. 


Last but not least, we do not explore scenarios involving more than two agents, as our primary focus is on studying several communication-related problems. By limiting the number of agents, we simplify the model. However, as shown in Figure \ref{FIG:1}, real-world applications often involve many entities. Therefore, tasks with significantly more than two agents warrant further investigation.

\section*{CRediT authorship contribution statement}
\textbf{Y.Wang:} Conceptualization, Methodology, Investigation
, Writing - Original Draft, Formal analysis, Visualization. \textbf{Y.Zhao:} Resources, Supervision, Data Curation, Writing - Original Draft, Writing - Review \& Editing, Project administration.

\section*{Declaration of competing interest}
The authors declare no potential conflicts of interest with respect to the research, authorship, and/or publication of this article.

\section*{Declaration of Generative AI and AI-assisted technologies in the writing process}
The authors declare that the AI-based language model, ChatGPT 4.0 by OpenAI was employed to enhance the spelling,
language, and grammar of this paper. After using the ChatGPT, the authors reviewed and edited the content as needed and take full responsibility for the content of the publication.


\section*{Supplementary material}
The following is the Supplementary material related to this article. \textbf{Movie S1}: The performance of a trained RL algorithm during testing in the ship cooperative navigation scenario, with N=1. \textbf{Movie S2}: The performance of a trained RL algorithm during testing in the ship cooperative navigation scenario, with N=2. \textbf{Movie S3}: The performance of a trained RL algorithm during testing in the ship cooperative collision avoidance scenario. The initial positions of give way ship and head on ship are randomly placed.

\section*{Appendix A. Main particulars of our USV}
We list main parameters of the USV used in our RL environment in Table \ref{tabel2}.
\begin{center}
    \captionof{table}{Main particulars of the USV}\label{tabel2}
    \begin{tabular}[h！]{p{5.2cm} p{2.0cm}}
        \toprule
        \textbf{Parameter} & \textbf{Value} \\
        \midrule
        Ship length & 7.0 ($m$)\\
        Ship breadth & 1.3 ($m$)\\
        Ship draught & 0.46 ($m$)\\
        Displacement volume of ship & 3.27 ($m^3$)\\
        Longitudinal coordinate of center of
gravity of ship & 0.25 ($m$)\\
        Rudder span length & 0.35 ($m$)\\
        Rudder area & 0.054 ($m^2$)\\
        Propeller diameter & 0.4 ($m^3$)\\
        Block coefficient & 0.55 ($m$)\\
        Propulsion coefficient & 0.6($m$)\\
        \bottomrule
    \end{tabular}
\end{center}

\bibliographystyle{cas-model2-names}
\bibliography{cas-refs}

\newpage
\end{multicols}
\end{document}